\pdfoutput=1

\documentclass[11pt]{article}

\usepackage[table]{xcolor}

\usepackage[preprint]{acl}
\usepackage{amsmath}

\usepackage{times}
\usepackage{latexsym}
\usepackage{multirow}
\usepackage{booktabs}
\usepackage{graphicx}
\usepackage{subcaption}
\usepackage{algorithm}
\usepackage{algorithmic}
\usepackage{amssymb}

\usepackage[T1]{fontenc}

\usepackage[utf8]{inputenc}

\usepackage{microtype}

\usepackage{inconsolata}

\usepackage{graphicx}

\title{CrossFormer: Cross-Segment Semantic Fusion for Document Segmentation}

\author{Tongke Ni\textsuperscript{1}, \ Yang Fan\textsuperscript{1}, \ Junru Zhou\textsuperscript{2}, \ Xiangping Wu\textsuperscript{1}, \ Qingcai Chen\textsuperscript{1}\\
\textsuperscript{1}Harbin Institute of Technology (Shenzhen), Shenzhen, China\\
\textsuperscript{2}Tencent Inc. Guangzhou, China\\
\texttt{\{nee\}@tanknee.cn} \ \texttt{\{yfan\}@stu.hit.edu.cn} \ \texttt{\{doodlezhou\}@tencent.com}\\
\texttt{\{wuxiangping,qingcai.chen\}@hit.edu.cn}\\
}

\begin{document}
\maketitle
\begin{abstract}

Text semantic segmentation involves partitioning a document into multiple paragraphs with continuous semantics based on the subject matter, contextual information, and document structure. Traditional approaches have typically relied on preprocessing documents into segments to address input length constraints, resulting in the loss of critical semantic information across segments. To address this, we present CrossFormer, a transformer-based model featuring a novel cross-segment fusion module that dynamically models latent semantic dependencies across document segments, substantially elevating segmentation accuracy. Additionally, CrossFormer can replace rule-based chunk methods within the Retrieval-Augmented Generation (RAG) system, producing more semantically coherent chunks that enhance its efficacy. Comprehensive evaluations confirm CrossFormer’s state-of-the-art performance on public text semantic segmentation datasets, alongside considerable gains on RAG benchmarks.

\end{abstract}

\section{Introduction}

\begin{figure*}[t]
  \centering
  \includegraphics[width=1\textwidth]{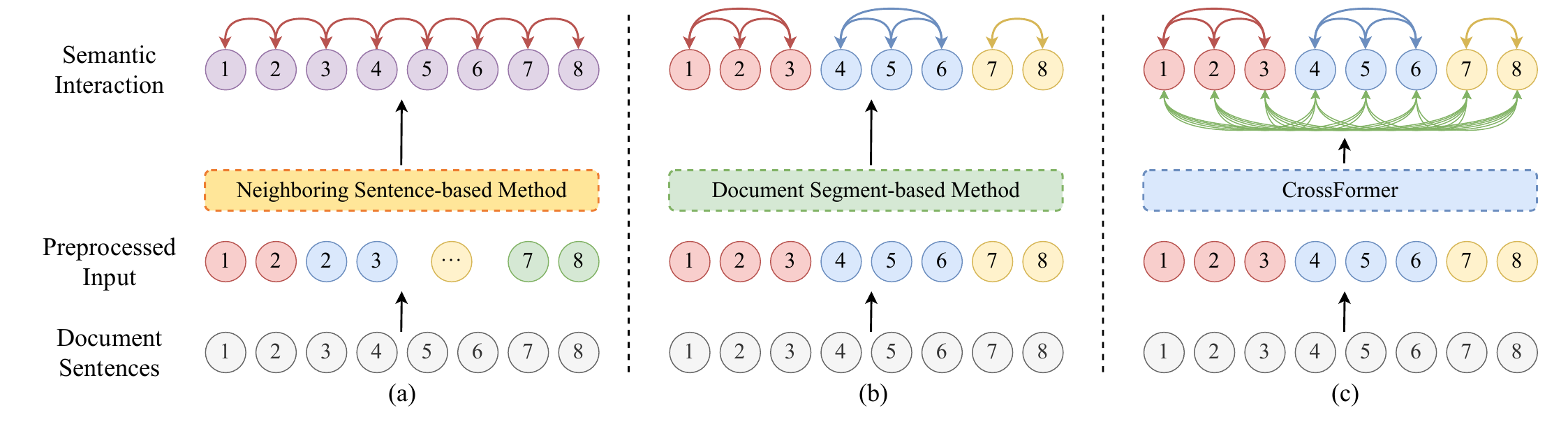}
  \caption{1a illustrates methods that leverages neighboring sentences of the candidate segmentation boundary to harness the contextual information \cite{lukasik2020textsegmentationcrosssegment}. 1b presents approaches that divide the document into segments, followed by the intra-segment semantic interaction. 1c introduces our proposed CrossFormer featuring CSFM to extract cross-segment semantic interaction depicted by green lines.}
  \label{fig:main}
\end{figure*}

Text semantic segmentation constitutes a fundamental challenge in the field of document analysis, focusing on the automated partition of documents into contiguous semantic units exhibiting thematic or contextual coherence \cite{hearst-1994-multi}. This task serves as a foundational preprocessing step for downstream applications spanning document summarization \cite{DBLP:conf/emnlp/XiaoC19,kupiec1995trainable}, structured information extraction \cite{chinchor1993evaluating}, and Retrieval-Augmented Generation (RAG) pipelines \cite{DBLP:journals/corr/abs-2404-10981}. Current methodologies bifurcate into multiple operational paradigms, including hierarchical segmentation \cite{DBLP:conf/lrec/BayomiL18,DBLP:conf/coling/HazemDSKC20} and linear segmentation \cite{DBLP:journals/coling/Hearst97}. 
Within the latter paradigm, CrossFormer focuses on the planar non-overlapping text semantic segmentation task.

Text semantic segmentation necessitates that sentences within the same paragraph cohesively revolve around a central topic, while maintaining minimal semantic overlap between distinct paragraphs. Early unsupervised methodologies identified segmentation boundaries through Bayesian models \cite{chen-etal-2009-global, riedl-biemann-2012-topictiling} or graph-based methods, where sentences were treated as nodes \cite{glavas-etal-2016-unsupervised}. On the other hand, supervised methods have leveraged pretrained language models (PLMs) derived from extensive corpora, subsequently fine-tuning them on annotated text semantic segmentation datasets. This paradigm enables a more comprehensive utilization of document content, yielding superior segmentation results \cite{DBLP:conf/naacl/KoshorekCMRB18,lukasik2020textsegmentationcrosssegment,zhang2021sequencemodelselfadaptivesliding,yu2023improvinglongdocumenttopic}.

However, the inherent constraints of transformer-based models, particularly their maximum context length, have led some approaches to partition documents into fixed-length segments \cite{yu2023improvinglongdocumenttopic}. This strategy, while pragmatic, introduces notable limitations. Chief among these is the absence of inter-segment correlations, which impedes the model's ability to capture sentence-level information spanning multiple segments. Consequently, the model's capacity to comprehend the broader semantic structure of the document is compromised.

To address this challenge, as illustrated in Figure \ref{fig:main}, we propose CrossFormer, a novel model incorporating CSFM, which models interactions between document segments, thereby enhancing segmentation performance by integrating cross-segment dependencies and facilitating a more robust understanding of the document’s overarching semantic hierarchy.

Leveraging CrossFormer’s capacity for document-level semantic segmentation, we propose its integration into Retrieval-Augmented Generation (RAG) systems. RAG has emerged as a critical framework for mitigating limitations inherent to large language models (LLMs), including temporal data obsolescence, domain-specific or proprietary data scarcity, and hallucination \cite{DBLP:journals/jmlr/IzacardLLHPSDJRG23, DBLP:journals/tacl/RamLDMSLS23}. A conventional RAG pipeline comprises retrieval and generation modules, where the retrieval stage involves chunk segmentation, vectorization, and query matching \cite{DBLP:journals/corr/abs-2312-10997}. 
Current RAG implementations often employ rule-based \cite{Chase2022LangChain} or LLM-driven chunking methods \cite{DBLP:conf/emnlp/DuarteMGF0O24, MetaChunking}. However, rule-based approaches frequently fail to preserve intra-chunk semantic coherence. And LLM-based segmentation methods, though effective in aligning chunks with semantic boundaries \cite{DBLP:conf/emnlp/DuarteMGF0O24, MetaChunking}, incur substantial latency. 
On this basis, we advocate for embedding CrossFormer as the chunk splitter within RAG, which can generate contextually coherent chunks while maintaining computational efficiency, thereby addressing the dual challenges of semantic fidelity and processing speed.

The main contributions of our paper are as follows:
\begin{itemize}
    \item We investigate the cross-segment semantic information loss caused by document preprocessing and propose a Cross-Segment Fusion Module (CSFM) to explicitly model latent semantic dependencies across document segments, thereby improve the performance of text semantic segmentation.
    \item We propose a novel model named CrossFormer featuring CSFM, which achieves state-of-the-art performance on the text semantic segmentation benchmarks. Ablation studies have demonstrated the effectiveness of the CSFM module.
    \item We integrate CrossFormer as the text chunk splitter into the RAG system and achieve superior performance compared to existing methods.
\end{itemize}

\section{Related Work}

\subsection{Text Semantic Segmentation}

Text semantic segmentation includes supervised and unsupervised methods. Unsupervised methods, like the Bayesian approach by \cite{chen-etal-2009-global}, use probabilistic generative models for document segmentation; \cite{riedl-biemann-2012-topictiling} applied a Bayesian method based on LDA, identifying segmentation boundaries through drops in coherence scores between adjacent sentences; \cite{glavas-etal-2016-unsupervised} introduced an unsupervised graph method, where sentences are nodes and edges represent semantic similarity, segmenting text by finding the maximum graph between adjacent sentences. Deep learning-based supervised methods have improved text semantic segmentation accuracy.  \cite{koshorek2018textsegmentationsupervisedlearning} modeled document segmentation as a sequence labeling task, predicting segmentation boundaries at separators, and introduced the WIKI-727k dataset, where supervised methods outperformed unsupervised ones. \cite{lukasik2020textsegmentationcrosssegment} introduced hierarchical BERT and other models to capture inter-sentence features, enhancing the performance of pre-trained language models fine-tuned on the WIKI-727k dataset.\cite{zhang2021sequencemodelselfadaptivesliding} proposed a model named SeqModel by integrating phonetic information and a sliding window to improve performance and lower resource consumption, also providing the Wiki-zh dataset for multilingual text semantic segmentation. \cite{yu2023improvinglongdocumenttopic, somasundaran2020two, arnold2019sector} utilizes explicit coherence modeling to enhance text semantic segmentation performance. Unsupervised methods lack the capability to iteratively refine and optimize their performance by labeled datasets. While current supervised learning approaches have enhanced performance through techniques such as phone embedding\cite{zhang2021sequencemodelselfadaptivesliding} and Contrastive Semantic Similarity Learning\cite{yu2023improvinglongdocumenttopic}, we contend that greater focus should be directed towards addressing the issue of cross-segment information loss induced by preprocessing strategies that partition long documents into fixed-length segments.

\subsection{RAG Text Chunk Splitter}

Text chunking serves as a foundational component of Retrieval-Augmented Generation (RAG) systems. Current methodologies bifurcate into rule-based and LLM-based paradigms. Rule-based approaches—such as chunking by sentences, paragraphs, or chapters—prioritize structural fidelity. Prior work by \cite{Chase2022LangChain} further combines separators, word counts, and character limits to define chunk boundaries. While these strategies preserve document structure, they inherently neglect semantic relationships during chunking.
LLM-based methods, conversely, emphasize semantic coherence. For instance, \cite{Kamradt2024SemanticChunking} cluster text via embeddings to aggregate semantically related content, while \cite{DBLP:conf/emnlp/DuarteMGF0O24} employ question-answering prompts to guide LLM-driven chunking for narrative texts. \cite{MetaChunking} advances this by computing sentence-level perplexity and deploying margin sampling to infer semantically informed chunk boundaries. However, LLM-based techniques face practical constraints: their parameterized architectures and autoregressive inference mechanisms incur substantial latency.
To balance chunk accuracy and computational cost, we require a lightweight model capable of accurately chunk documents according to semantics.

\section{Methodology}

This section elucidates the architecture of CrossFormer and its integration within the Retrieval-Augmented Generation (RAG). The exposition commences with a formal definition of the text semantic segmentation task and the preprocessing methodology of documents. Subsequently, we introduce the proposed Cross-Segment Fusion Module (CSFM), which enhances semantic coherence across document segments. Finally, we provide a comprehensive exposition of CrossFormer's system integration within the RAG pipeline, emphasizing its capacity to generate semantic chunks while demonstrating enhancements in answer generation.

\subsection{CrossFormer}\label{ch:3.1}

\begin{figure*}[htbp]
    \centering
    \begin{subfigure}[b]{0.8\textwidth}
        \includegraphics[width=1\textwidth]{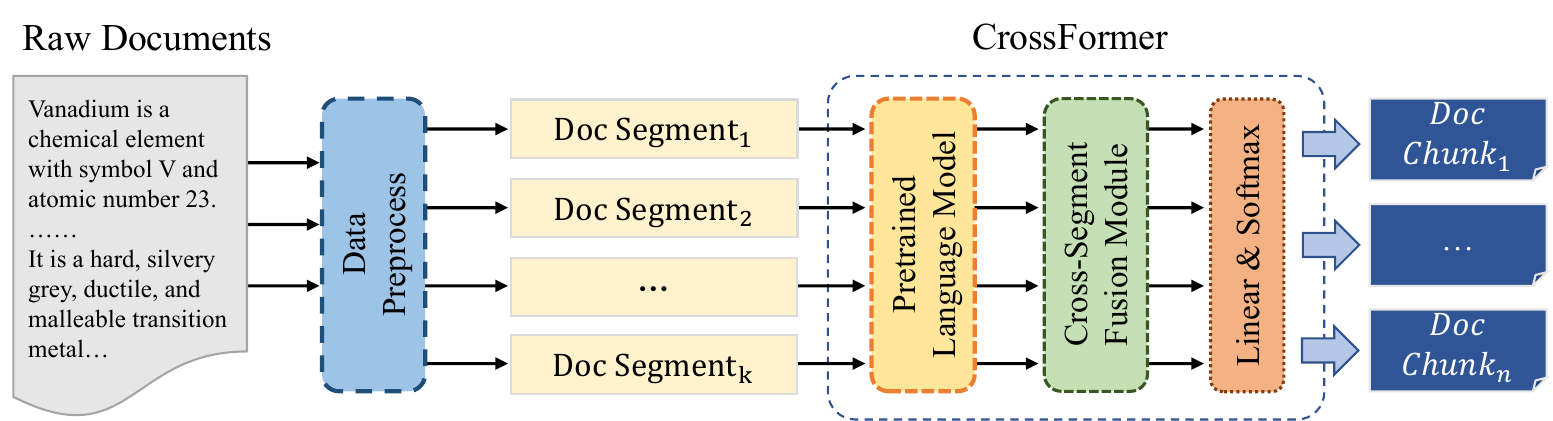}
        \caption{CrossFormer Pipeline}
        \label{fig:cross-former-process-flow}
    \end{subfigure}
    \begin{subfigure}[b]{0.4\textwidth}
        \includegraphics[width=1\textwidth]{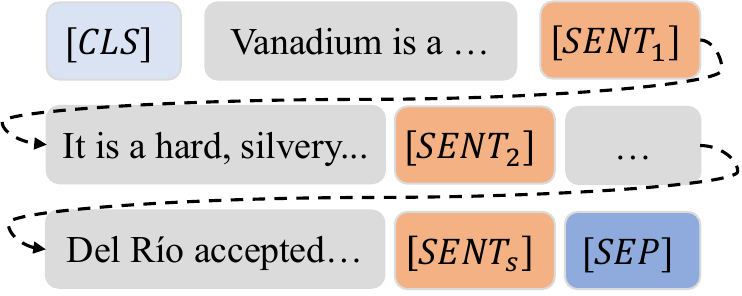}
        \caption{Document Segment Example}
        \label{fig:document-segment-example}
    \end{subfigure}
    \hspace{5pt}
    \begin{subfigure}[b]{0.4\textwidth}
        \includegraphics[width=1\textwidth]{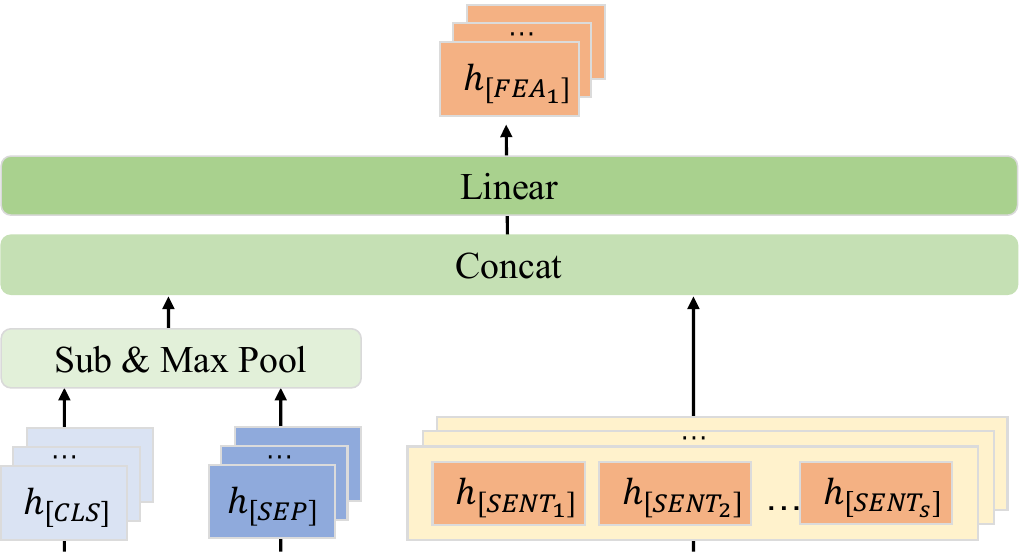}
        \caption{Cross-Segment Fusion Module}
        \label{fig:cross-segment-fusion-module}
    \end{subfigure}
    \caption{Architecture of CrossFormer. \ref{fig:cross-former-process-flow} illustrates the pipeline of CrossFormer for text semantic segmentation task and its architecture, which consists of a pre-trained language model, Cross-Segment Fusion Module (CSFM), and a linear classifier. \ref{fig:document-segment-example} shows an example of a preprocessed document segment as input to the model. \ref{fig:cross-segment-fusion-module} demonstrates the detailed structure of CSFM.}
    \label{fig:crossformer-flow}
\end{figure*}

\textbf{Task Definition}\quad We frame the task of text semantic segmentation as a sentence-level sequential labeling problem \cite{lukasik2020textsegmentationcrosssegment,koshorek2018textsegmentationsupervisedlearning,zhang2021sequencemodelselfadaptivesliding,yu2023improvinglongdocumenttopic}. Consider a document \(\mathcal{D} = \{s_1, s_2, \ldots, s_n\}\) comprising \(n\) sentences, where \(s_i\) denotes $i$-th sentence. A binary segmentation label \(y_i \in \{0, 1\}\) is assigned to each sentence \(s_i\), such that \(y_i = 1\) denotes the terminal boundary of a paragraph unified by semantic coherence, while \(y_i = 0\) signifies continuity within the same topical segment. The objective is to train a function \(f: \mathcal{D} \rightarrow \{0, 1\}^n\) capable of predicting \(y_i\) for each \(s_i\) according to the context.  
We append a special separator $[SENT]$ to the end of each sentence \(s_i\), yielding a modified document \(\mathcal{D}' = \{s'_1, s'_2, \ldots, s'_n\}\). The model subsequently evaluates whether the \([SENT]\) token appended to \(s'_i\) demarcates a segment boundary. Formally, let \(\mathcal{T} = \{T_1, T_2, \ldots, T_m\}\) represent the complete set of semantic paragraphs within \(\mathcal{D}\), where:  
\begin{align}
\mathcal{D} = \bigcup_{i=1}^m T_i \quad \text{and} \quad T_i \cap T_{i'} = \varnothing \;\; \forall i \neq i'.
\end{align}
Each paragraph \(T_i = \{s^{(i)}_1, s^{(i)}_2, \ldots, s^{(i)}_j, \ldots\}\) constitutes an ordered sequence of sentences sharing a cohesive semantic topic. The segmentation task thus reduces to partitioning \(\mathcal{D}\) into the set \(\mathcal{T}\) of disjoint, topically homogeneous partitions \(T_i\), achieved through precise identification of boundary-inducing \([SENT]\) tokens.

\textbf{Document pre-processing}\quad Considering the excessive length of documents typically involved in text semantic segmentation tasks, it is essential to adopt appropriate methodologies for effectively modeling long documents.
In accordance with established long-text modeling techniques, we integrate truncation and segmentation approaches \cite{dong2023surveylongtextmodeling}. As shown in Algorithm \ref{alg:docproc}, given a document $\mathcal{D}$ with length $N$. We first segment $\mathcal{D}$ according to the separators specified by the task (e.g., line breaks or periods). Denote the set of sentences obtained as $\{s_i\}_{i = 1}^{n}$, where $|s_i|\leq L$ for all $i = 1,\cdots,n$ and $\sum_{i = 1}^{n}|s_i|=N$, $L$ is the maximum length of a single sentence. Then, we concatenate these sentences in sequence to form $k$ document segments $\{d_j\}_{j = 1}^{k}$, such that $\sum_{j = 1}^{k}|d_j| = N$ and $|d_j|\leq M$ for all $j = 1,\cdots,k$ and $k<K$, where $M$ is the maximum length of a document segment. The parameters $L$ and $M$ are hyperparameters with $L< M$. The parameter $K$ represents the maximum number of segments. A larger $K$ allows the model to capture longer-range semantic information, but it also increases the demand on GPU memory. These document segments are then assembled into a batch and fed into the model for training or inference. When generating document segments, the special tokens $[CLS]$ and $[SEP]$ will be appended at the beginning and end respectively.

\textbf{Cross-Segment Fusion Module (CSFM)} Once a document is splitted into multiple segments, these segments are compiled into a batch. As depicted in Figure \ref{fig:crossformer-flow}, we propose CSFM that integrates global and sentence-level information for classification.
\begin{align}
    h_{\text{seg}}&=h_{\text{[CLS]}}-h_{\text{[SEP]}}\label{eq:topic-sub1}\\
    h_{\text{global}}&=\max([h_{\text{seg-1}},h_{\text{seg-2}},...,h_{\text{seg-k}}]) \label{eq:topic-sub2}
\end{align}

In order to extract the semantic information of segments, we select the pre-trained special tokens $[CLS]$ and $[SEP]$ in the transformer model. We construct $h_{\text{seg}}$ through formula \eqref{eq:topic-sub1}. Thus, we can obtain the semantic representation $h_{\text{seg-1}},h_{\text{seg-2}},\cdots,h_{\text{seg-k}}$ of k segments. Then, we apply max-pooling to obtain the largest semantic component from these representations and acquire $h_{\text{global}}$ to represent the global semantic information of all segments.
\begin{align}
    h_{\text{concat}}&=\text{Concat}(h_{\text{global}}, h_{\text{[SENT]}})\label{eq:ffn1}\\
    h_{\text{[FEA]}}&=\text{Linear}_2(\text{Linear}_1(h_{\text{concat}}))\label{eq:ffn2}
\end{align}

Upon the global semantic embedding \( h_{\text{global}} \) is computed, we concatenate it with each separator embedding \( h_{[SENT]} \). Subsequently, two linear layers are applied to the concatenated vector $h_{\text{concat}}$ to get $h_{\text{[FEA]}}$, the representations of separators integrated with global semantic information. This process is mathematically encapsulated in equation \eqref{eq:ffn1} and \eqref{eq:ffn2}. 

Following the acquisition of the output from the CSFM, \( h_{\text{[FEA]}} \) is input into the Linear layer and applied the Softmax function to get the classification results. Finally, We segment the document into semantic paragraphs according to the classification results .

\begin{algorithm}
    \small
    \caption{Document Processing for Text Semantic Segmentation}\label{alg:docproc}
    \begin{algorithmic}[1]
    \STATE \textbf{Input:} A document \( \mathcal{D} \) consisting of \( S \) sentences. Maximum sentence length \( L \), maximum document segment length \( M \). Maximum number of document segments $K$
    \STATE \textbf{Output:} Batch \(B\) of document segments for model \(\mathcal{M}\)
    \STATE \(\{s_1,s_2,\ldots,s_n\} \gets\) Split and Truncate \(\mathcal{D}\) using \(\text{separators}\) such that \(\max(|s_i|)\leq L\) for \(i = 1,\ldots,n\)
    \STATE Initialize an empty list \(B\)
    \STATE \(j\gets1\)
    \STATE \(segment\gets\) empty string
    \FOR{\(i = 1\) to \(n\)}
        \IF{\(|segment|+|s_i|+2>M\)}
            \STATE \(segment\gets segment + [SEP]\)
            \STATE Add \(segment\) to \(B\)
            \STATE \(segment\gets [CLS] + s_i + [SENT]\)
            \STATE \(j\gets j + 1\)
        \ELSE
            \STATE \(segment\gets segment + s_i + [SENT]\)
        \ENDIF
    \ENDFOR
    \IF{\(segment\) is not empty}
        \STATE Add \(segment + [SEP]\) to \(B\)
    \ENDIF
    \IF{\(j>K\)}
        \STATE \(B \gets \{segment_1, \dots, segment_K\}\)
    \ENDIF
    \RETURN \(B\)
    \end{algorithmic}
\end{algorithm}

During the training phase, the CrossFormer is trained using a standard cross-entropy loss function to maximize the likelihood of the training data, thereby enhancing the model's ability to accurately identify segmentation separators within the text sequence.

\subsection{CrossFormer as RAG Text Chunk Splitter}

\begin{figure}[t]
  \centering
  \includegraphics[width=\columnwidth]{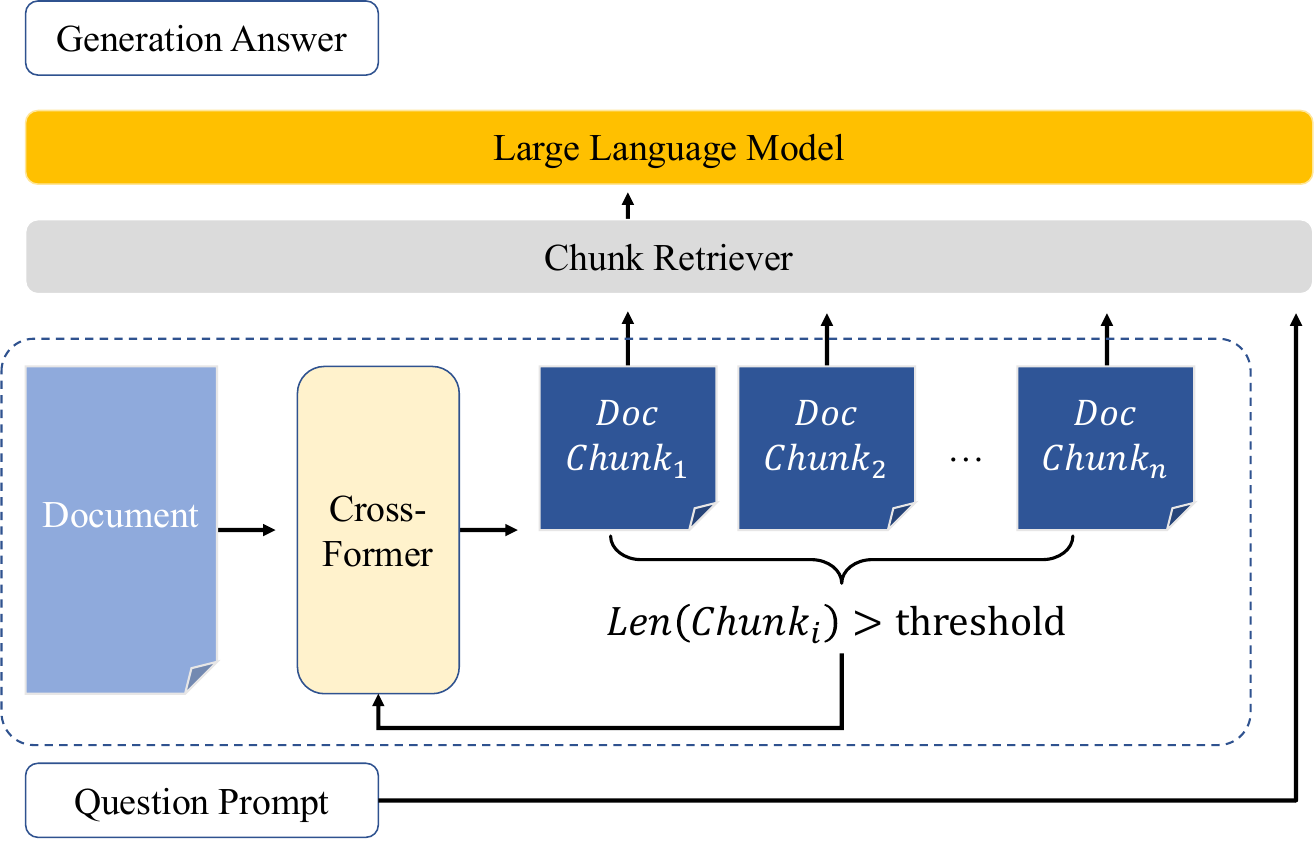}
  \caption{A flowchart depicting the integration of CrossFormer into the RAG system as a text chunk splitter.The process begins with document input, followed by CrossFormer-based chunking. The top-k semantically relevant chunks are retrieved, and a large language model generates the final answer using the retrieved chunks.}
  \label{fig:seprm-flow}
\end{figure}

We integrate CrossFormer as the text chunk splitter of the RAG system to get better semantic chunks.
As illustrated in Figure \ref{fig:seprm-flow}, we have devised a systematic process for splitting long documents. Initially, the CrossFormer is utilized to split the input document, yielding a sequence of text chunks. Subsequently, the length of each segment is evaluated. When the length surpass a predefined threshold, the text chunk will be input into the splitting queue for further recursive processing until the model determines that no additional segmentation is required or the text chunk length falls below the specified threshold. Upon achieving the segmented text chunks, a retriever and a question prompt are employed to conduct relevance retrieval, generating the context which is subsequently input into the LLMs to get the final answer.

\section{Experiment}

\subsection{Text Semantic Segmentation}

\textbf{Experimental Setup}\quad We performed testing and evaluation on several widely used text semantic segmentation datasets, including the English datasets WIKI-727k and WIKI-50 \cite{koshorek2018textsegmentationsupervisedlearning}, and the Chinese dataset WIKI-zh \cite{zhang2021sequencemodelselfadaptivesliding}. The CrossFormer was trained on the training subsets of these datasets and subsequently evaluated on their respective test subsets, with an inference batch size of 2.

\textbf{Dataset statistic}\quad The WIKI-727k dataset is a large-scale collection derived from Wikipedia. All images, tables, and other non-textual elements have been removed, and the documents have been tokenized into sentences using the PUNKT tokenizer from NLTK \cite{Bird2006NLTKTN}. Semantic segmentation of the documents was executed in accordance with the table of contents. For the experiments, the training set provided by WIKI-727k was utilized to train the model, and its performance was assessed on the corresponding test set to ensure a fair comparison with other models. The WIKI-50 dataset, a subset derived from WIKI-727k, was employed for faster evaluation of model performance. WIKI-zh, constructed similarly to WIKI-727k but utilizing data from the Chinese Wikipedia, constitutes a larger dataset. The statistical characteristics of these three datasets are presented in Table \ref{tab:dataset-statistic}, which illustrates that WIKI-727k in English contains a greater number of sentences, a comparable number of topics, yet fewer sentences per topic. The statistics of WIKI-50 exhibit a close resemblance to those of WIKI-727k.

\setlength{\tabcolsep}{3pt}
\begin{table}
  \centering
  \small
  \begin{tabular}{lccccc}
        \toprule
        Dataset & Docs & S/Doc & SLen & T/Doc & S/T \\
        \midrule
        WIKI-727k & 727,746 & 52.65 & 129.20 & 6.18 & 8.52 \\
        WIKI-zh & 820,773 & 11.77 & 43.28 & 5.57 & 2.11 \\
        \midrule
        WIKI-50 & 50 & 61.40 & 125.10 & 7.68 & 7.99 \\
        \bottomrule
  \end{tabular}
  \caption{\label{tab:dataset-statistic}
    Statistics of text semantic segmentation datasets. Docs denotes the total number of documents. S/Doc and T/Doc denote the average number of sentences and topics per document, respectively. SLen denotes the average number of characters per sentence. S/T denotes the average number of sentences per topic.
  }
\end{table}

\textbf{Model training settings}\quad As discussed in Chapter \ref{ch:3.1}, to effectively model long documents within a limited context, we developed a scheme that integrates truncation and segmentation. In the experimental phase, we adapt the training and inference parameters for various models and devices. For models such as BERT \cite{Devlin2019BERTPO} and RoBERTa \cite{Liu2019RoBERTaAR}, which have a context length of 512 tokens, we established the maximum length \(M\) of document segments at 512 tokens and did not impose a limit on the maximum number of document segments $K$. For the Longformer \cite{Beltagy2020LongformerTL} model, which supports a longer context (up to 4096), we set the maximum number of document segments to either 4 or 8, contingent upon the available GPU memory, and accordingly adjusted the maximum length of each document segment.

\subsection{RAG Text Chunk Splitter}

\setlength{\tabcolsep}{2pt}
\begin{table}
    \centering
    \small
    \begin{tabular}{lcccc}
        \toprule
        Dataset & ID & Source & Metric & \#Data \\
        \midrule
        \emph{Single-Document QA} & & & & \\
        \midrule
        NarrativeQA & 1-1 & Literature, Film & F1 & 200 \\
        Qasper & 1-2 & Science & F1 & 200 \\
        MultiFieldQA-en & 1-3 & Multi-field & F1 & 150 \\
        \rowcolor{green!20} MultiFieldQA-zh & 1-4 & Multi-field & F1 & 200 \\
        \midrule
        \emph{Multi-Document QA} & & & & \\
        \midrule
        HotpotQA & 2-1 & Wikipedia & F1 & 200 \\
        2WikiMultihoopQA & 2-2 & Wikipedia & F1 & 200 \\
        MuSiQue & 2-3 & Wikipedia & F1 & 200 \\
        \rowcolor{green!20} DuReader & 2-4 & Baidu Search & Rouge-L & 200 \\
        \bottomrule
    \end{tabular}
    \caption{Overview of the LongBench dataset \cite{bai-etal-2024-longbench}. Chinese datasets are highlighted in light green; the Source column indicates the provenance of these datasets, and the \#Data column specifies the number of samples contained within each dataset.}\label{tab:longbench-statistic}
\end{table}

\textbf{Experiment Setup}\quad To explore whether replacing the non-semantic chunking methods with a text semantic segmentation model can bring end-to-end performance improvements to the RAG system, we followed \cite{bai-etal-2024-longbench}, selecting the same eight datasets, including single-document datasets NarrativeQA, MultiFieldQA-en, Qasper, MultiFieldQA-zh, and multi-document datasets HotpotQA, MuSiQue, 2WikiMultihopQA, DuReader. Notably, the MultiFieldQA-zh and DuReader datasets are Chinese datasets, while the others are in English. In the experiment, we use CrossFormer:Roberta trained on the WIKI-727k training set as the text chunk splitter.

\textbf{Preprocessing and RAG Workflow} In these datasets of LongBench \cite{bai-etal-2024-longbench}, the original documents do not have ready-made sentence segmentation results, and the addition style of separators varies across different datasets. Therefore, we processed the original text, for example, by adding line breaks after Chinese and English periods, and only at the end of sentences, to preserve sentence integrity as much as possible. Then, we split the original text into sentences based on single line breaks. After obtaining the sentences, we constructed batches according to Algorithm \ref{alg:docproc}, and input them into CrossFormer to get the semantic segmentation results. Following the process illustrated in Figure \ref{fig:seprm-flow}, we split the long documents into text chunks, and use bge-m3 \cite{Chen2024BGEMM} embedding model to convert both the text chunks and the question prompt into embeddings. We calculated the cosine similarity scores between the question embedding and the text chunk embeddings, selected the $\emph{K Chunks}$ as context, filled the preset prompt template in LongBench, and then input it into LLMs for answer generation.

\section{Evaluation}

\subsection{Results of Text Semantic Segmentation}

\defcitealias{lukasik2020textsegmentationcrosssegment}{1}
\defcitealias{zhang2021sequencemodelselfadaptivesliding}{2}
\defcitealias{yu2023improvinglongdocumenttopic}{3}

\setlength{\tabcolsep}{1pt}
\begin{table}[t]
  \centering
  \small
  \begin{tabular}{lccc}
    \toprule
    \multirow{2}{*}{Model}         & \multicolumn{3}{c}{WIKI-727k}       \\
    ~    & P & R & $F_1$ \\
    \midrule
    BERT-Base + Bi-LSTM \citeyearpar{lukasik2020textsegmentationcrosssegment}  & 67.3 & 53.9 & 59.9 \\
    Hier. BERT (24-Layers) \citeyearpar{lukasik2020textsegmentationcrosssegment} & 69.8 & 63.5 & 66.5 \\
    Cross-segment BERT-L 128 \citeyearpar{lukasik2020textsegmentationcrosssegment} & 69.1 & 63.2 & 66.0 \\
    Cross-segment BERT-L 256 \citeyearpar{lukasik2020textsegmentationcrosssegment} & 61.5 & 73.9 & 67.1  \\
    SeqModel:BERT-Base \citeyearpar{zhang2021sequencemodelselfadaptivesliding} & 70.6 & 65.9 & 68.2 \\
    Longformer-Base(TSSP+CSSL) \citeyearpar{yu2023improvinglongdocumenttopic} & - & - & 77.16 \\
    Longformer-Large(TSSP+CSSL)$^{\dagger}$ & \textbf{82.40} & 74.81 & 78.42 \\
    \midrule
    CrossFormer:Longformer-Base(ours) & 81.33 & 71.57 & 76.14 \\
    CrossFormer:Longformer-Large(ours) & 82.02 & \textbf{75.98} & \textbf{78.88} \\
    \bottomrule
  \end{tabular}
  \caption{\label{tab:wiki-727k-performance}
    Performance results on the WIKI-727k test set. Bold numbers indicate the best performance. ${\dagger}$ denotes our reproduced results. $P$ means Precision, $R$ means Recall.
  }
\end{table}

The performance of CrossFormer on the WIKI-727k dataset is presented in Table \ref{tab:wiki-727k-performance}. We evaluate the performance of CrossFormer on the WIKI-727k test set using both the Longformer-Base and Longformer-Large models for training. Compared to other text semantic segmentation models, CrossFormer attains the highest F1 score.

\setlength{\tabcolsep}{3pt}
\begin{table}
  \centering
  \small
  \begin{tabular}{lccc}
    \toprule
    \multirow{2}{*}{Model}     & \multicolumn{3}{c}{WIKI-zh}       \\
    ~    & P & R & $F_1$ \\
    \midrule
    Cross-segment BERT-Base 128-128 &   61.2 & \textbf{80.2} & 69.4 \\
    SeqModel:BERT-Base & 78.4 & 69.5 & 73.7 \\
    SeqModel:StructBERT-Base & \textbf{79.2} & 72.7 & 75.8 \\
    SeqModel:RoBERTa-Base & 74.6 & 73.7 & 74.2 \\
    SeqModel:ELECTRA-Base & 73.5 & 76.6 & 75.0 \\
    \midrule
    CrossFormer:RoBERTa-Base (ours) & 78.01 & 78.60 & \textbf{78.31} \\
    \bottomrule
  \end{tabular}
  \caption{\label{tab:wiki-zh-performance}
    The evaluation results on the WIKI-zh test set, Bold numbers indicate the best performance. $P$ means Precision, $R$ means Recall.
  }
\end{table}

The evaluation results on the Wiki-zh dataset are presented in Table \ref{tab:wiki-zh-performance}. We utilized the RoBERTa \cite{Liu2019RoBERTaAR} for this experiment. The experimental findings reveal that CrossFormer achieved superior performance compared to other leading methods, with a notable improvement in the $F_1$ metric by 2.51. These results suggest a significant enhancement in performance.

\subsection{Results on RAG benchmark}

Eight datasets from the LongBench were employed to evaluate retrieval and context compression. The performance of CrossFormer as a chunk splitter is presented in Table \ref{tab:longbench-rag-performance}. The experiment was on the number of input chunks generated by different chunk splitter methods. We also explored the impact of the number of input chunks on the performance.

Moreover, the performance indicate that CrossFormer demonstrates superior performance on English datasets in comparison to other methodologies. Conversely, its performance on Chinese datasets is merely average, which may be attributed to the utilization of only the English version of the Wiki-727k corpus during the training phase.

In terms of performance metrics across each dataset, CrossFormer outperformed other methods in most settings, thereby demonstrating its effectiveness. Although Lumber has achieved good results in some settings, due to its use of LLMs for chunking, it has a certain disadvantage in terms of speed. However the LC-C and LC-R methods have good performance on some datasets, the chunk method based on CrossFormer has achieved better results on more datasets after integrating semantic information.

\subsection{Ablation Study}

\setlength{\tabcolsep}{3pt}
\begin{table}
  \centering
  \small
  \begin{tabular}{lccc}
    \toprule
    \multirow{2}{*}{CrossFormer Base Model}      & \multicolumn{3}{c}{WIKI-727k}       \\
    ~    & Precision & Recall & $F_1$ \\
    \midrule
    BERT-Base & 67.16 & 73.42 & 70.15 \\
    RoBERTa-Base & 73.96 & 72.66 & 73.31 \\
    Longformer-Base & 81.33 & 71.57 & 76.14 \\
    Longformer-Large & \textbf{82.02} & \textbf{75.98} & \textbf{78.88} \\
    \bottomrule
  \end{tabular}
  \caption{\label{wiki-727k-model-compare}
    Performance on the Wiki-727K test set from our proposed CrossFormer based on different PLMs. Bold numbers denote the best performance.
  }
\end{table}

\setlength{\tabcolsep}{3pt}
\begin{table}
  \centering
  \small
  \begin{tabular}{lccc}
    \toprule
    \multirow{2}{*}{CrossFormer Base Model}         & \multicolumn{3}{c}{WIKI-727k}       \\
    ~    & Precision & Recall & $F_1$ \\
    \midrule
    Longformer-Base & 80.05& \textbf{71.87}& \textbf{75.74}\\
    \quad w/o CSFM & \textbf{80.43}& 70.68& 75.24\\
    \midrule
    Longformer-Large & \textbf{82.02}& \textbf{75.98}& \textbf{78.88}\\
    \quad w/o CSFM & 80.53& 75.31& 77.83\\
    \midrule
    BERT-Base & 67.16 & \textbf{73.42}& \textbf{70.15}\\
    \quad w/o CSFM & \textbf{76.92}& 63.6& 69.63 \\
    \bottomrule
  \end{tabular}
  \caption{\label{wiki-727k-ablation}
    Ablation study for CSFM on WIKI-727k test set. Bold numbers denote the best performance.
  }
\end{table}

\begin{table*}
  \centering
  \small
    \begin{tabular}{lccccccccccc}
        \toprule
        \multirow{2}{*}{Method} & \multicolumn{4}{c}{\textbf{Single-Doc QA}} & \multicolumn{4}{c}{\textbf{Multi-Doc QA}} &  \multirow{2}{*}{Avg} & \multirow{2}{*}{En-Avg} & \multirow{2}{*}{Zh-Avg} \\
        \cmidrule(lr){2-5} \cmidrule(lr){6-9}
        & 1-1 & 1-2 &	1-3 & 1-4$^\star$ &2-1 & 2-2 &	2-3 & 2-4$^\star$ &  &  & \\
        \midrule
          w/o retrieval & 17.72 & 44.67 & 50.14 & 60.82 & 46.36 & 45.59 & 23.66 & 29.68 & 39.83 & 38.02 & 45.25 \\
        \midrule
        \emph{3 Chunks} \\
        \midrule
        LC-C& 22.65 & 43.08 & \textbf{49.39} & 62.51 & \textbf{52.06} & 40.18 & 27.06 & \textbf{28.52} & 40.68 & 39.07 & 45.52 \\
        LC-R & 20.32 & 40.16 & 48.27 & 59.12 & 51 & 42.19 & \textbf{28.17} & 28.51 & 39.72 & 38.35 & 43.82 \\
        Lumber& 19.95 & \textbf{43.99} & 48.34 & 61.98 & 51.87 & 41.89 & 24.27 & 28.29 & 40.07 & 38.39 & 45.14 \\
        \rowcolor{gray!40} CrossFormer & \textbf{23.5} & 40.36 & 47.25 & \textbf{63.61} & 51.24 & \textbf{47.65} & 26.25 & \textbf{28.52} & \textbf{41.05} & \textbf{39.38} & \textbf{46.07} \\
        \midrule
        \emph{5 Chunks} \\
        \midrule
        LC-C & 22.28 & \textbf{42.14} & 48.22 & 62.66 & \textbf{53.32} & 47.67 & 29.08 & 28.68 & 41.76 & 40.45 & 45.67 \\
        LC-R & 21.08 & 40.66 & 48.64 & 61.35 & 53.19 & 45.78 & 25.42 & 28.56 & 40.59 & 39.13 & 44.96 \\
        Lumber& 20.94 & 41.64 & \textbf{49.75} & \textbf{62.88} & 52.31 & 44.95 & 25.43 & \textbf{29.59} & 40.94 & 39.17 & \textbf{46.24} \\
        \rowcolor{gray!40} CrossFormer & \textbf{23.08} & 41.44 & 46.96 & 62.26 & 52.5 & \textbf{52.2} & \textbf{34.39} & 27.62 & \textbf{42.56} & \textbf{41.76} & 44.94  \\
        \bottomrule
    \end{tabular}
  \caption{\label{tab:longbench-rag-performance}
    Experiment results on LongBench dataset. The rows marked in gray indicate our proposed CrossFormer, LC-C denotes Langchain's \textit{CharacterTextSplitter}, LC-R denotes Langchain's \textit{RecursiveCharacterTextSplitter}, and Lumber indicates the LumberChunk \cite{DBLP:conf/emnlp/DuarteMGF0O24} method with Qwen2.5-7B-Instruct Model. $\emph{K Chunks}$ denotes the number of the most relevant chunks selected as context.$\star$ indicates the Chinese dataset. The name of datasets corresponding to each ID are shown in Table \ref{tab:longbench-statistic}.
  }
\end{table*}

\begin{figure}[t]
  \centering
  \includegraphics[width=\columnwidth]{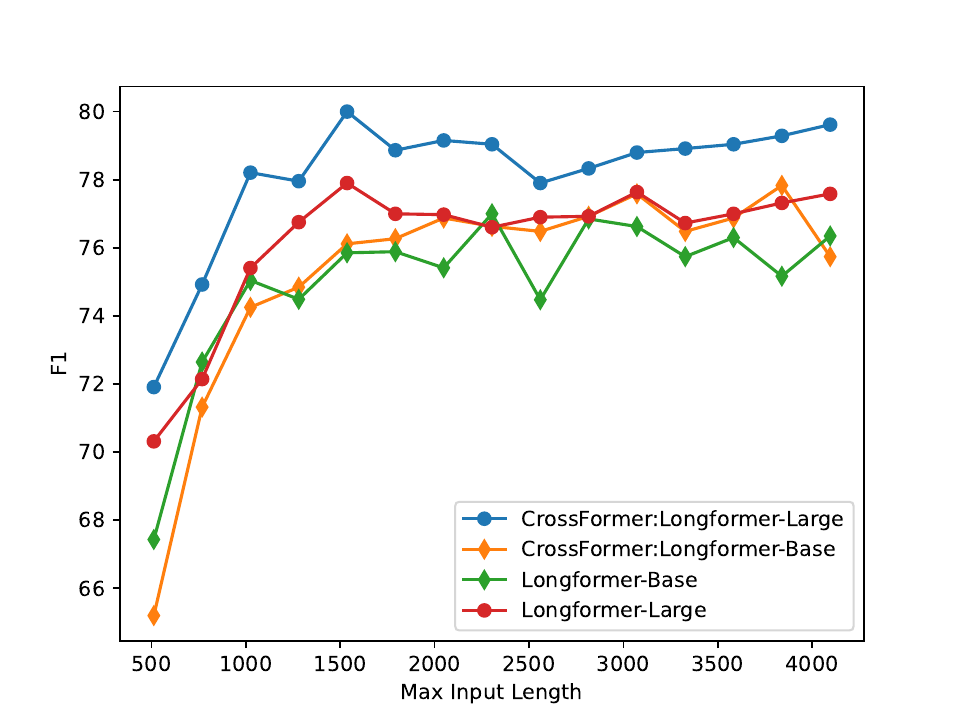}
  \caption{The influence of max input length of CrossFormer on the WIKI-50 dataset \cite{koshorek2018textsegmentationsupervisedlearning}. Models without the prefix "CrossFormer" are ablation experiments that do not contain CSFM.}
  \label{fig:input-length-effect}
\end{figure}

To validate the effectiveness of the proposed CrossFormer method, ablation experiments were conducted on the WIKI-727k test set utilizing various models and the proposed CSFM. As illustrated in Table \ref{wiki-727k-model-compare}, the method was trained and evaluated on BERT-Base \cite{Devlin2019BERTPO}, RoBERTa-Base \cite{Liu2019RoBERTaAR}, Longformer-Base \cite{Beltagy2020LongformerTL}, and Longformer-Large \cite{Beltagy2020LongformerTL} models. The results indicate that the CrossFormer method achieves the highest performance with Longformer, likely due to its capacity to model longer contexts at the model level, thereby enabling the inclusion of more sentences within the same document segment. This performance advantage aligns with expectations. Additionally, the performance variations observed between the Longformer-Base and Longformer-Large models confirm that an increase in model parameters results in substantial performance improvements.

As illustrated in Table \ref{wiki-727k-ablation}, a series of comprehensive ablation experiments were conducted on the WIKI-727k test set. The results indicate that the module achieved performance enhancements in both Recall and F1 metrics. Additionally, ablation experiments were performed across different models, demonstrating that the CSFM can generally augment the performance. In addition, as depicted in Figure \ref{fig:input-length-effect}, experiments were also carried out on the CrossFormer with different input lengths on the WIKI-50 dataset. The experimental results revealed that the CrossFormer incorporating CSFM exhibited improved performance in most input lengths compared to the model without CSFM. Secondly, it was discovered that when the input length was less than approximately 1500, the performance improved with the input length. When the length was greater than approximately 1500, the performance improvement it brought was relatively minor.

\section{Conclusion}

In this paper, we introduce a new text semantic segmentation model, termed CrossFormer, which incorporates a novel Cross-Segment Fusion Module (CSFM). This module is designed to deal with the cross-segment information loss between document segments by constructing document segment embedding and global document embedding to alleviate the context information loss when predicting segmentation boundaries. Furthermore, we integrate the CrossFormer into the RAG system as a semantic chunk splitter for documents, aiming to overcome the limitations of previous rule-based methods that fail to adequately leverage semantic information within the text. Empirical results indicate that our proposed CrossFormer achieves state-of-the-art performance across several public datasets. Additionally, the model demonstrates improvements in RAG performance.

\section{Limitations}

Despite its innovative Cross-Segment Fusion Module and strong performance in text semantic segmentation, CrossFormer has limitations. Since CrossFormer is a deep learning method, when it is used as a text chunk splitter, its running speed is slower than rule-based methods but faster than LLM-based methods. Secondly, CrossFormer cannot precisely control the upper limit of the length of text chunks. Therefore, it may be necessary to combine rule-based methods to output an appropriate length. Finally, as a linear text semantic segmentation model, CrossFormer cannot output partially overlapping text chunks, which is required in some scenarios of the RAG system.

\bibliography{custom}

\appendix

\end{document}